\documentclass[10pt,twocolumn,letterpaper]{article}

\usepackage{iccv}
\usepackage{times}
\usepackage{epsfig}
\usepackage{siunitx}
\usepackage{booktabs}
\usepackage{graphicx}
\usepackage{amsmath}
\usepackage{amssymb}
\usepackage{bbm}
\usepackage{algorithm}
\usepackage{algpseudocode}
\usepackage{multirow}
\usepackage{caption}
\usepackage{subcaption}
\usepackage[table,xcdraw]{xcolor}

\makeatletter
\NewDocumentCommand{\LeftComment}{s m}{%
  \Statex \IfBooleanF{#1}{\hspace*{\ALG@thistlm}}\(\triangleright\) #2}
\makeatother
\usepackage[pagebackref=true,breaklinks=true,letterpaper=true,colorlinks,bookmarks=false]{hyperref}

\iccvfinalcopy 

\def\iccvPaperID{1735} 

\ificcvfinal\fi

\begin{document}

\title{Sample Less, Learn More: Efficient Action Recognition via Frame Feature Restoration}

\author{Harry Cheng\\
Shandong University\\
Qingdao, China\\
{\tt\small xaCheng1996@gmail.com}
\and
Yangyang Guo\\
National University of Singapore\\
Singapore, Singapore\\
{\tt\small guoyang.eric@gmail.com}
\and
Liqiang Nie\\
Harbin Institute of Technology\\
Shenzhen, China\\
{\tt\small nieliqiang@gmail.com}
\and
Zhiyong Cheng\\
Shandong Artificial Intelligence Institute\\
Jinan, China\\
{\tt\small jason.zy.cheng@gmail.com}
\and
Mohan Kankanhalli\\
National University of Singapore\\
Singapore, Singapore\\
{\tt\small mohan@comp.nus.edu.sg}
}

\maketitle
\ificcvfinal\thispagestyle{empty}\fi
\begin{abstract}
Training an effective video action recognition model poses significant computational challenges, particularly under limited resource budgets. Current methods primarily aim to either reduce model size or utilize pre-trained models, limiting their adaptability to various backbone architectures. This paper investigates the issue of over-sampled frames, a prevalent problem in many approaches yet it has received relatively little attention. Despite the use of fewer frames being a potential solution, this approach often results in a substantial decline in performance. To address this issue, we propose a novel method to restore the intermediate features for two sparsely sampled and adjacent video frames. This feature restoration technique brings a negligible increase in computational requirements compared to resource-intensive image encoders, such as ViT. To evaluate the effectiveness of our method, we conduct extensive experiments on four public datasets, including Kinetics-400, ActivityNet, UCF-101, and HMDB-51. With the integration of our method, the efficiency of three commonly used baselines has been improved by over 50\%, with a mere 0.5\% reduction in recognition accuracy. In addition, our method also surprisingly helps improve the generalization ability of the models under zero-shot settings.
\end{abstract}

\section{Introduction}
The ready availability of millions of videos~\cite{Kinectics, UCF101, HMDB, ActNet} and billion-scale model parameters~\cite{MAE_B, VIT_22B} put video action recognition at a substantial computational bottleneck. Early attempts to address this issue have mainly been devoted to reducing model size by integrating temporal modules~\cite{TDN, TEA, STM} into 2D visual encoders, as opposed to relying on their computationally intensive 3D counterparts. Recently, the rise of vision-language pre-training~\cite{CLIP,BLIP, BLIP2, vqa-tip} has spurred researchers to investigate the potential of large multi-modal pre-trained models~\cite{CLIP, Transfer_VL_ko}. A commonly used approach is to fine-tune pre-trained weights~\cite{ActionClip, Transfer_VL_ko} and mitigate the burden of training from scratch. While these advancements have yielded encouraging outcomes, their generalizability may be limited by certain factors, such as specific backbones.

\begin{figure}[t]
    \centering
    \includegraphics[width=0.45\textwidth]{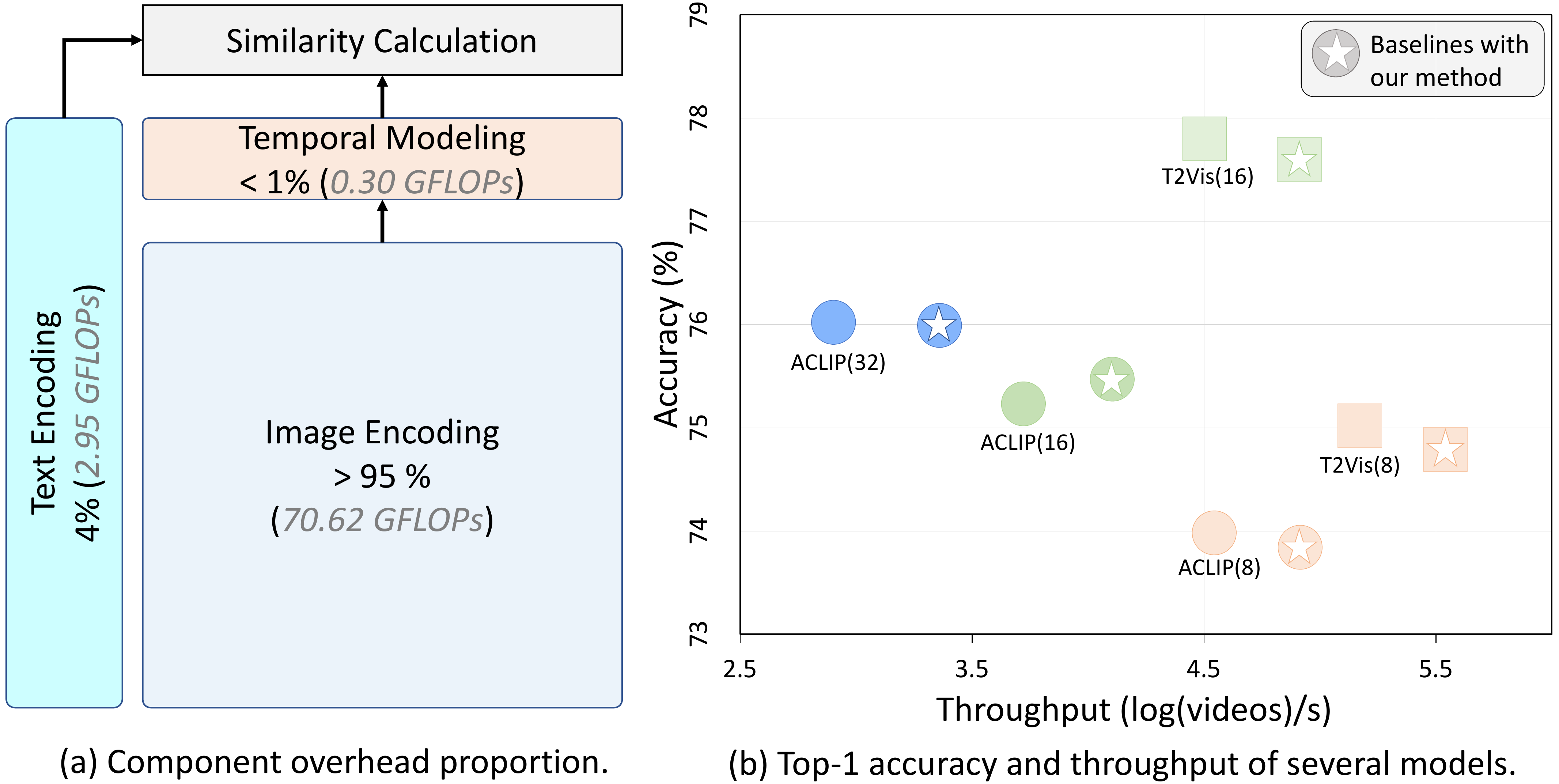}
    \caption{
    (a) The computational costs of the different components in ActionCLIP~\cite{ActionClip} with 16 sampled frames, where the image encoding takes the majority of overhead.
    (b) Model performance on Kinetics-400~\cite{Kinectics}. We use shapes and colors to differentiate backbones and the sample frame number, respectively. Our method achieves a better efficiency-effectiveness trade-off.
    }
    \label{fig:Time_Per_Plot}
    \vspace{-1em}
\end{figure}

This paper throws light on dense frame sampling of videos, a topic that the existing literature has largely overlooked. Figure~\ref{fig:Time_Per_Plot}a shows that the image encoding module accounts for the bulk of overhead in a typical model. As a result, reducing the number of sampled video frames significantly helps bypass the computationally expensive image encoding. The model training can thus be accelerated considerably. Furthermore, given the indispensable role of frame sampling in diverse action recognition paradigms, this approach can also alleviate reliance on specific backbone models and contribute to reducing the carbon footprint of video processing.

\begin{figure}[t]
    \centering
    \includegraphics[width=0.45\textwidth]{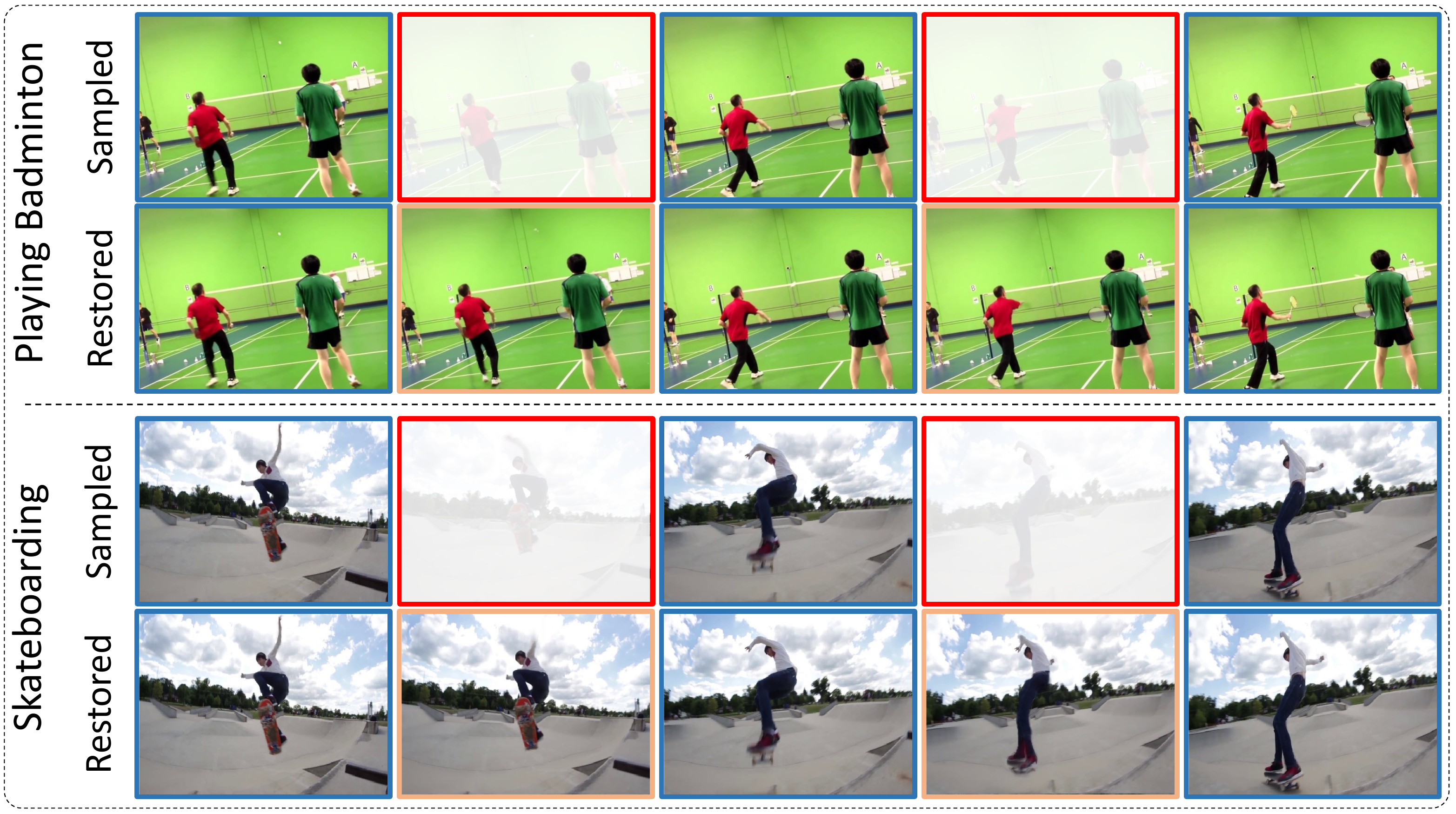}
    \caption{Frame restoration result illustration. It can be seen that the restored frames (orange border) maintain both spatial and temporal features of the original videos.}
    \label{fig:Restore_performance}
    \vspace{-1em}
\end{figure}

However, using fewer frames often comes at the cost of drastic performance drops. These results can be primarily attributed to the loss of context that provides discriminative cues for action recognition. To address this issue, this paper presents a novel learning scheme, \textbf{Sample Less Learn More (SLLM)}, to enhance the limited context without introducing heavier computations. In particular, SLLM reconstructs the intermediate vision features rather than encoding them from scratch. Two supporting points constitute the foundation of our approach: i) Compared to complete image encoding, generating intermediate features significantly reduces computational costs; and ii) recent advancements in frame drop restoration~\cite{restoration, MM_res, JiangInterpolation2018, ZhouInterpolation2021}, which generate damaged frames during video transmission, have demonstrated the feasibility of generating high-quality results based on limited context (refer to Figure~\ref{fig:Restore_performance}).

\begin{figure*}[t]
    \centering
    \includegraphics[width=0.75\textwidth]{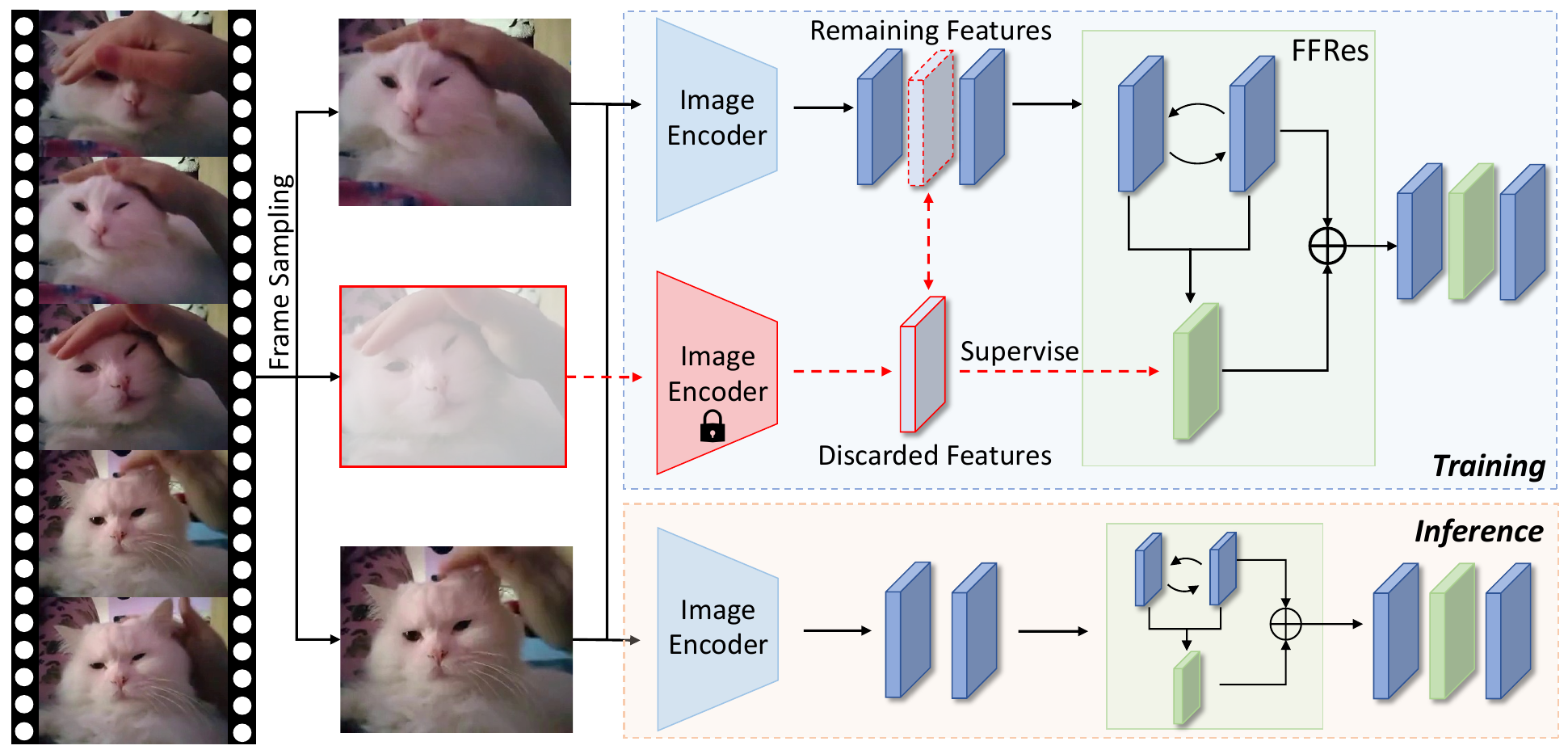}
    \vspace{-1em}
    \caption{
        Overview of our SLLM.
        In the training stage, we employ the FFRes module to restore the discarded features based on the features of the remaining frames.
        The discarded frames provide supervision for feature restoration using a frozen image encoder.
        During inference, we exclude frame discard operation for more efficiency.}    
    \label{fig:Overview}
    \vspace{-1em}
\end{figure*}

We present an overview of the proposed SLLM in Figure~\ref{fig:Overview}. 
Compared to conventional frameworks that use complete sampled frames (\textbf{sample more}), SLLM is distinguished by its efficient approach of using fewer frames (\textbf{sample less}).
In particular, we first discard a fraction of sampled frames.
Thereafter, the vision encoder receives the remaining frames as input, and the Frame Feature Restoration (FFRes) module is utilized to restore the intermediate features corresponding to the discarded frames (\textbf{learn more}).
We train the FFRes module to align with the features extracted from a vision encoder with frozen weights.
During inference, we directly restore the features based on the remaining surrounding frames without any supervision. 
In addition, we also introduce a pre-trained captioner to enrich the action label with diverse semantics. With the augmented external knowledge, the labels are characterized by increased inter-label discrimination. 

We conduct extensive experiments to evaluate the effectiveness of our method on four widely-used datasets: Kinetics-400~\cite{Kinectics}, ActivityNet~\cite{ActNet}, UCF-101~\cite{UCF101}, and HMDB-51~\cite{HMDB}. 
The experimental results demonstrate that our method contributes significantly to improving efficiency, resulting in an increase of over 50\% in both GFLOPs and video throughput. 
Furthermore, as shown in Figure~\ref{fig:Time_Per_Plot}b, SLLM maintains the original performance of the baselines with only a marginal decrease in accuracy of 0.5\% when compared to the original baselines.
The results also indicate that our approach improves the generalizability of the baselines, as reflected in a notable 3\% increase in accuracy under the zero-shot setting.

The main contributions of this work are three-fold:
\begin{itemize}
    \item To the best of our knowledge, we are the first to improve the efficiency of action recognition by reducing the workload of image encoding while maintaining the performance. Our approach is not constrained to specific backbones and can be seamlessly applied to various action recognition models.
    \item We devise a simple yet effective multi-modal learning scheme to generate the intermediate frame features. Besides improving action recognition efficiency, it potentially benefits other  tasks with sample frames, such as video deepfake detection.
    \item The experimental results demonstrate that our approach significantly enhances efficiency while exhibiting performance on par with several SOTA models. Furthermore, our proposed approach effectively improves the models' generalizability under the zero-shot setting.
\end{itemize}

\section{Related Work}
\subsection{Video Action Recognition}
Video action recognition has long been a popular research topic over the past few years~\cite{frame_active, Regularization_smooth, Regularization_Multigrid}. Early endeavors focus on the development of backbones, such as 3D-CNNs~\cite{I3D, S3D, C3D} and Vision Transformer (ViT)~\cite{ViT, VideoSwin}. 
followed by later studies on data augmentation~\cite{DataAug_Act, DTR, Regu_1} and sampling strategies~\cite{OCSampler, MGSampler}. Despite advanced recognition accuracy being achieved, the efficiency issue has grown into a significant concern for researchers recently.

To tackle this problem, several prior attempts have been dedicated to optimizing the models' parameter size. To this end, temporal modules~\cite{TA2N, TSN} are developed to replace 3D-CNNs for temporal context modeling. TSM~\cite{TSM} utilizes a temporal shift module to replace convolutions, and Wang et al.~\cite{TDN} proposed a two-level difference modeling paradigm to capture local and global motions. 
Besides offering benefits to CNN-based models~\cite{liu2023hs}, these approaches also boost the development of ViT-based models such as ViViT~\cite{vivit}, TimeSFormer~\cite{Timesformer}, and MViT~\cite{MViT, MviTV2}. 
Unlike these methods, recent approaches have benefited mainly from large vision language pre-training~\cite{BLIP, BLIP2}. 
According to this new paradigm, visual frames and textual labels are simultaneously trained in a dual-stream network~\cite{Two_stream} by similarity matching. 
As the majority of parameters of these methods require only fine-tuning or remain frozen, the computation demand is thereby decreased.
For instance, ActionCLIP~\cite{ActionClip} utilizes the pre-trained CLIP~\cite{CLIP} to align videos and actions by the contrastive learning objective. 
Text2Vis~\cite{Transfer_VL_ko} incorporates an additional frozen action feature matrix in the similarity calculation process, leading to acceleration of parameter update and thus improving model efficiency.

Although the aforementioned approaches show favorable results in improving efficiency, they may be restricted by specific backbones or require complex model adaptations. Unlike these approaches, our method, which aims to reduce the sampled frames, presents the best of both worlds: i) Our method considerably enhances model efficiency via video frame reduction and is agnostic to different backbones; ii) The model performance is largely maintained, with a decline of no more than 0.5\%.

\subsection{Video Frame Restoration}
Video frame restoration aims to recover lost frames or increase the frame rate~\cite{KimRestoration2018, HeRestoration2020} from a continuous video stream. The existing work can be categorized into uni-modal and multi-modal methods based upon the input modality. In the former, visual information is solely utilized to generate the missing frames. Specifically, Jiang et al.~\cite{JiangInterpolation2018} synthesized intermediate frames via estimating optical flows, ALANET~\cite{GuptaInterpolation2020} employ a combination of self-attention and cross-attention modules between consecutive frames to generate an optimized representation for each frame. Regarding multi-modal ones~\cite{MM_res, vqa-tomm}, Cheng et al.~\cite{restoration} utilized the entire audio track to restore video frames, where an alignment module is employed to synchronize the video and audio streams. AHFVR~\cite{WeiRestoration2022} employs the hierarchical fine-grained representation structure to repair the visual frames based on the latent correlation between low-level features from the audio and haptic signals. 
The recovered images can preserve most of the spatial and temporal information of the original video~\cite{GuptaInterpolation2020, JinRestoration2020}, and some examples are shown in Figure~\ref{fig:Restore_performance}. 
Pertaining to our method, we generate the intermediate vision features rather than decoding the missing frame as these restoration approaches do. As a result, our approach introduces minor incremental time overhead to the model (as discussed in Section~\ref{Sec:time_comp}).

\section{Methodology}

\subsection{Preliminary of Paradigms}
\label{Sec:prel}
Current video action recognition methods mostly follow two training paradigms: classification-based~\cite{TanClassification2021} and matching-based~\cite{JuMatching22}. We illustrate the matching-based paradigm in the following as an example. Models that adhere to the paradigm typically consist of a vision feature extractor $g(\cdot)$ and an action identifier $f_s(\cdot)$. Let $\mathcal{D} = \{(\mathcal{V}i, y_i)\}_{i=1}^{N}$ denote a dataset comprising $N$ videos, and each video $\mathcal{V}_i$ is associated with a label $y_i$ from $M$ candidate action labels. During training, the goal is to find the optimal model parameters $\theta \in \Theta$, where $\Theta$ represents the continuous parameter space. To achieve this, the expected loss over the training data is defined to minimize the empirical risk,
\begin{equation}
    R_\mathcal{D}(\theta):=\frac{1}{N} \sum_{i=1}^N \mathcal{L}_{sim} \left(\theta, f_s(g(\mathcal{I}_i), \mathbf{S})\right),
\label{Eqn:risk_simi}
\end{equation}
where $\mathcal{I}_i = \{\mathbf{I}_j\}_{j=1}^{T}$ represents the $T$ frames sampled from $\mathcal{V}_i$, and 
$\mathbf{S} = [\mathbf{s}_1, \mathbf{s}_2, \cdots, \mathbf{s}_M]$ is an $M$-row matrix, wherein each $\mathbf{s}_k \in \mathbf{S}$ represents the embedding of a specific action label; The function $f_s(\cdot)$ measures the similarity between the sampled frame features $g(\mathcal{I}_i)$ and each $\mathbf{s}_k$. The model selects the highest similarity candidate as the predicted action with respect to the input video. The $\mathcal{L}_{sim} $ is the contrastive loss function to pull the matching video and action embedding $\bar{\mathbf{s}}$ closer and update $\theta$,
\begin{equation}
\mathcal{L}_{sim} =-\left(\log \frac{\exp \left(f_s(g(\mathcal{I}_i), \bar{\mathbf{s}})\right)}{\sum_{k=1}^M \exp \left(f_s(g(\mathcal{I}_i), \mathbf{s}_k)\right)}\right).
\label{EQN: simi_loss}
\end{equation}

Regarding the classification-based paradigm, a classifier $f_c:g(\mathcal{I}_i) \rightarrow \mathbb{R}^{M}$ is utilized to map the visual features to predicted labels and the risk during training is similarly formalized as,
\begin{equation}
R_\mathcal{D}(\theta):=\frac{1}{N} \sum_{i=1}^N \mathcal{L}_{cls} \left(\theta, f_c(g(\mathcal{I}_i)), y_i\right).
\label{Eqn:risk_cls}
\end{equation}

\subsection{Sample Less Learn More}
\subsubsection{Sampled Frame Reduction (Sample Less)}

Section~\ref{Sec:prel} demonstrates the indispensability of the image encoding $g(\mathcal{I}_i)$ in both classification and matching-based paradigms. 
However, this operation often results in significant computational costs. In light of this challenge, we endeavor to circumvent this problem by reducing the number of sampled frames. 
Specifically, we first conduct frame sampling using the original model strategy and subsequently downsize the frame set $\mathcal{I}_i$ based on a sampling filter $r$,
\begin{equation}
{\mathcal{I}}_i^{\prime} = \{\mathbf{I}_j \in \mathcal{I}_i:j \equiv 1\ (\mathrm{mod}\ r)\},
\label{Eqn:discarding}
\end{equation}
where ${\mathcal{I}}_i^{\prime}$ represents the remaining frame set and $r > 1$ is an integer that controls the number of frames. For example, downsizing $\mathcal{I}_i$ by half can be achieved with $r=2$. $\mathcal{I}_i^{\prime}$ is then fed into the vision feature extractor, where the empirical risk is modified as follows:
\begin{equation}
    \hat{R}_\mathcal{D}(\theta):=\frac{1}{N} \sum_{i=1}^N \mathcal{L}_{sim} (\theta, f_s(g(\mathcal{I}_i^{\prime}), \mathbf{S})).
\end{equation}

By encoding only a subset of the sampled frames, the computational load of image encoding can be reduced. Nevertheless, this approach may result in a degradation of performance. To mitigate this issue, we propose a feature restoration module that approximates the features of the discarded frames.
\begin{figure}[t]
    \centering
    \includegraphics[width=0.48\textwidth]{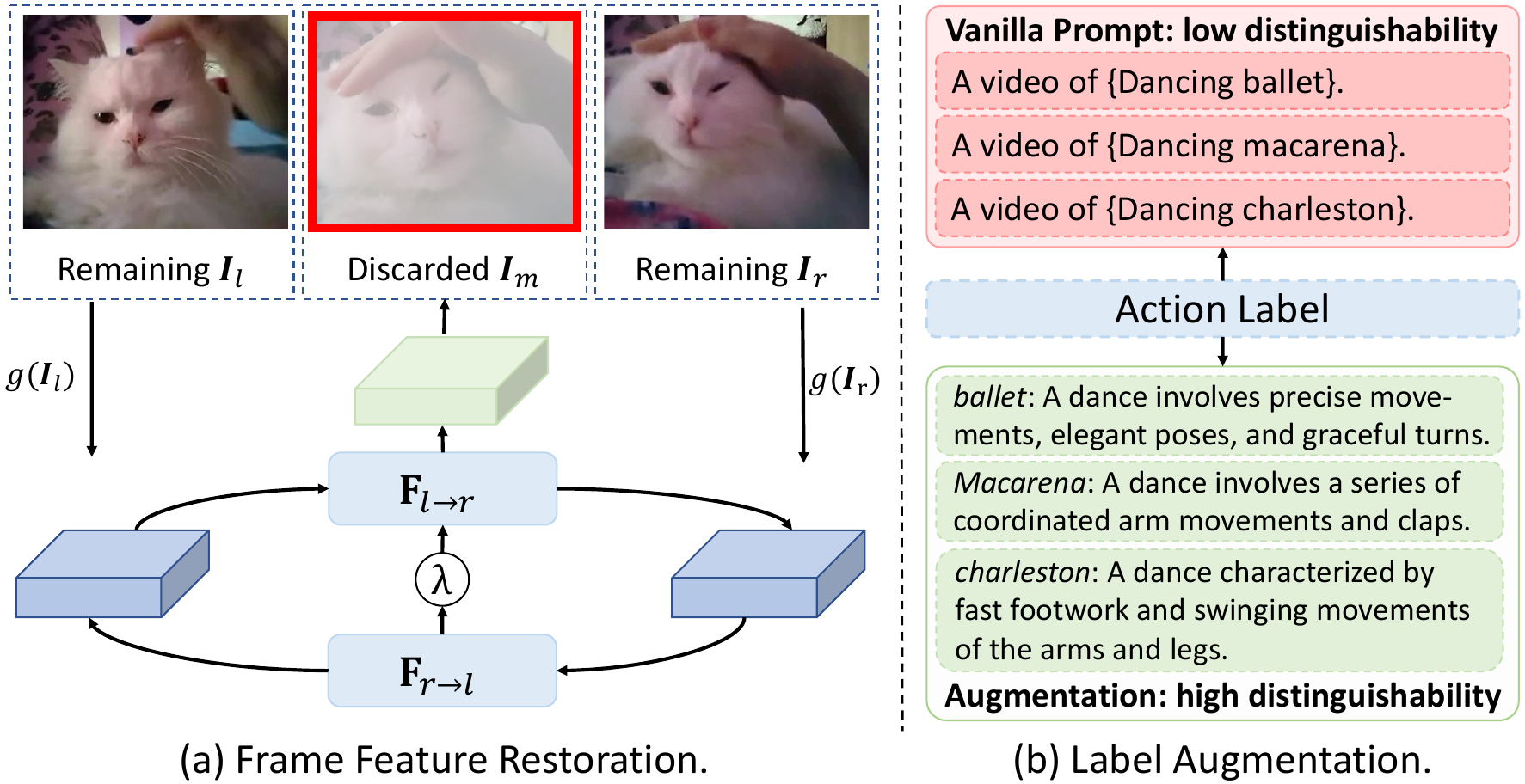}
    \caption{(a) Illustration of the feature restoration approach. We take the consecutive $\mathbf{I}_l$ and $\mathbf{I}_r$ as references to predict the discarded features. (b) We incorporate the external knowledge from a pre-trained captioner to enrich the action label with more semantic information.}
    \label{fig:Visual_Text_Branch}
\end{figure}
\subsubsection{Frame Feature Restoration (Learn More)}
Equation~\ref{Eqn:discarding} implies that the retained and discarded frames are interleaved in the original frame set $\mathcal{I}_i$, where each discarded frame is preceded by a leading frame and followed by a subsequent one from ${\mathcal{I}}_i^{\prime}$. 
To provide a concrete illustration of the FFRes module, we consider a specific example involving three frames, namely $\mathbf{I}_l$, $\mathbf{I}_m$, and $\mathbf{I}_r$. 
$\mathbf{I}_l$ and $\mathbf{I}_r$ represent consecutive retained frames, and $\mathbf{I}_m$ signifies a discarded frame located between them. 
The FFRes module restores the discarded features from the adjacent remaining ones, as shown in Figure~\ref{fig:Visual_Text_Branch}a, using the following equations:
\begin{equation}
\left\{\begin{array}{l}
    \hat{\mathbf{F}}_{l \rightarrow r}=h(g(\mathbf{I}_l), g(\mathbf{I}_r)), \\
    \hat{\mathbf{F}}_{r \rightarrow l}=h(g(\mathbf{I}_r), g(\mathbf{I}_l)), \\
    \hat{\mathbf{F}}_{m} = (1-\lambda) \times \hat{\mathbf{F}}_{l \rightarrow r} + \lambda \times \hat{\mathbf{F}}_{r \rightarrow l},
\end{array}\right.
\end{equation}
where $\hat{\mathbf{F}}_{m}$ denotes the predicted features of the discarded frame $\mathbf{I}_m$, and $\lambda = \frac{m-l}{r}$ determines the contribution of each adjacent frame in predicting the discarded features. The function $h(\cdot)$ can be formalized as:
\begin{equation}
    h(g(\mathbf{I}_l), g(\mathbf{I}_r)) = \operatorname{softmax}(\frac{\mathbf{g} \mathbf{g}^\intercal}{\sqrt{d_{\mathbf{g}}}})\mathbf{g},
\label{EQN:h}
\end{equation}
where $\mathbf{g} = [g(\mathbf{I}_l), g(\mathbf{I}_r)]$ denotes the feature vector of remaining frames. 
We leverage this operation to perform non-local learning between the two remaining frames~\cite{ActionClip}.
The restored features will be interpolated with the remaining ones for action prediction. For instance, in the matching-based paradigm, the prediction for action labels can be formalized as:
\begin{equation}
    \hat{y}_i = f_s(\{\cdots, g(\mathbf{I}_l), \hat{\mathbf{F}}_{m}, g(\mathbf{I}_r), \cdots\}, \mathbf{S}).
    \label{Eqn:new_feat}
\end{equation}

\begin{algorithm}[t]
\caption{Training and Inference with SLLM}  
\label{alg:restore}
\begin{algorithmic}
    \Require
      The original sample frames $\mathcal{I}_i$, the sampling filter $r$, the action embedding $\mathbf{S}$, the feature extractor $g(\cdot)$, the FFRes module $h(\cdot)$, and the action identifier $f_s(\cdot)$;
   \Ensure
      The predictive label $\hat{y}_i$;
    \State $\mathcal{I}_i^{\prime}$, $\overline{\mathcal{I}_i^{\prime}}$ $\gets \{\}$; \Comment{Initialize the remaining and discarded frame set}
    \For{$j \gets 1$ to $T$}
    \If{$j\!\!\mod r == 1$}
        \State $\mathcal{I}_i^{\prime} \gets \mathcal{I}_i^{\prime} \cup \{\mathbf{I}_j\}$;
    \Else
        \State $\overline{\mathcal{I}_i^{\prime}} \gets \overline{\mathcal{I}_i^{\prime}} \cup \{\mathbf{I}_j\}$;
    \EndIf
    \EndFor
    
    \State $g(\mathcal{I}_i^{\prime}) \gets \{g(\mathbf{I})|\mathbf{I} \in \mathcal{I}_i^{\prime}\}$;      \Comment{Remaining frame encoding}
    \State $\mathcal{F}$ $\gets \{\}$; \Comment{Initialize the restored feature set}
    \For {$\mathbf{I}_l, \mathbf{I}_r \in \mathcal{I}_i^{\prime}$}           \Comment{$\mathbf{I}_l$ and $\mathbf{I}_r$ are consecutive frames in $\mathcal{I}_i^{\prime}$}
        \State $\hat{\mathbf{F}}_{m} \gets (1-\lambda) \times h(g(\mathbf{I}_l), g(\mathbf{I}_r)) + \lambda \times h(g(\mathbf{I}_r), g(\mathbf{I}_l))$;
        \State $\mathcal{F} \gets \mathcal{F} \cup \hat{\mathbf{F}}_{m}$;
    \EndFor \\
    \If{Is\_Training}        \Comment{Training Stage}
        \For{$\hat{\mathbf{F}}_{m}, \mathbf{I}_m \in \mathcal{F}, \overline{\mathcal{I}_i^{\prime}}$}    \Comment{Supervision}
            \State $\operatorname{D}(\hat{\mathbf{F}}_{m} || g(\mathbf{I}_m)) \gets \sum g(\mathbf{I}_m)\log\frac{\hat{\mathbf{F}}_{m}}{g(\mathbf{I}_m)}$;  \Comment{Update FFRes}
        \EndFor
        \State \Return $\hat{y}_i \gets f_s(\{\cdots, g(\mathbf{I}_l), \hat{\mathbf{F}}_{m}, g(\mathbf{I}_r), \cdots\}, \mathbf{S})$;
    \Else                \Comment{Inference Stage}
        \State \Return $\hat{y}_i \gets f_s(\{\cdots, g(\mathbf{I}_l), \hat{\mathbf{F}}_{m}, g(\mathbf{I}_r), \cdots\}, \mathbf{S})$;
    \EndIf
            
\end{algorithmic}
\end{algorithm}

\subsubsection{Training Objectives}
Algorithm~\ref{alg:restore} outlines the key steps involved in training with SLLM. In a nutshell, we use the sampling filter $r$ to downsize the sampled frame set $\mathcal{I}_i$ and restore the discarded frames' features using the remaining frames $\mathcal{I}^{\prime}$.

\noindent\textbf{Training.} The objective of SLLM is to generate features that have the same distribution as the discarded frames based on a limited number of remaining frames, such that,
\begin{equation}
    \operatorname{D}(\hat{\mathbf{F}}_{m} || g(\mathbf{I}_m)) \approx 0,
\end{equation}
where the $g(\mathbf{I}_m)$ is the features extracted from discarded frames, and $\operatorname{D}$ denotes a metric used to measure the similarity between distributions of $\hat{\mathbf{F}}_{m}$ and $g(\mathbf{I}_m)$.

Specifically, we leverage a mutual information-based loss function in our implementation:
\begin{equation}
    \mathcal{L}_{res} = \sum g(\mathbf{I}_m)\log\frac{\hat{\mathbf{F}}_{m}}{g(\mathbf{I}_m)},
    \label{Eqn:KL}
\end{equation}
where the feature extraction $g(\mathbf{I}_m)$ is performed with a frozen image encoder involving no gradient update. We then combine $\mathcal{L}_{res}$ with the original base action recognition loss, e.g., $\mathcal{L}_{sim}$ or $\mathcal{L}_{cls}$, to update model parameters.

\noindent\textbf{Inference.}
During inference, we keep only the frame feature restoration and exclude the feature supervision from Equation~\ref{Eqn:KL}. In this way, we first obtain the vision features from the remaining frames using the vision encoder, and then generate the \textbf{pseudo} features of discarded frames for each of two adjacent remaining frames.
We finally predict the action labels based on the concatenation of these two sets of frame features.

\subsection{Action Label Augmentation} 
In addition to performing feature restoration, we also optimize the action label prompting, which has gained significant popularity recently~\cite{ActionClip}. 
Figure~\ref{fig:Visual_Text_Branch}b illustrates that prompts are typically presented in a fixed structure, such as \textit{A video of \{\}}, in common matching-based models~\cite{Transfer_VL_ko, JuMatching22}. 
However, some actions may require external knowledge to be correctly recognized, and are thus challenging for these fixed prompt templates. For instance, identifying Charleston or Salsa dance poses great difficulties for non-experts.
To address the issue, we propose leveraging prior knowledge, such as language models~\cite{ChatGPT}, to improve the diversity of label prompts. Specifically, we enrich the action label as
\begin{equation}
\hat{\mathbf{c}} \sim P_{LLM}(\mathbf{c}|a, inst, \tau),
\label{Eqn:aug}
\end{equation}
where $P_{LLM}(\cdot)$ refers to a language model, $\mathbf{c}$ represents the explanation sequence generated by $P_{LLM}(\cdot)$ pertaining to the action label $a$, $inst$ corresponds to the query for the sequence $\mathbf{c}$, such as `what does this action mean?'. We then choose a sub-sequence $\hat{\mathbf{c}}$ from $\mathbf{c}$ based on the temperature $\tau$ and take it as the caption. This process allows the captioner to interpret specialized terms, e.g., \textit{dancing macarena}, into meaningful sentences, such as \textit{a dance that involves a series of coordinated arm movements and claps.}

\begin{table}[]
\caption{Time complexity analysis between ActionCLIP and SLLM with sampling filter $r=2$. Our method mainly optimizes computationally intensive image encoding.}
\scalebox{0.70}{
\begin{tabular}{l|c|cc}
\toprule \midrule
\multirow{3}{*}{Model} & \multicolumn{1}{c|}{\multirow{3}{*}{Image Encoder}} & \multicolumn{2}{c}{FFRes Module}                                                     \\  \cmidrule(lr){3-4}   
                       & \multicolumn{1}{c|}{ }                       & \multicolumn{1}{c}{Supervison} & \multicolumn{1}{c}{\multirow{2}{*}{Restoration}} \\
                       & \multicolumn{1}{c|}{ }                       & (gradient-free)                & \multicolumn{1}{c}{}                             \\ \midrule
ActionCLIP             &  $T \times \mathcal{T}_{img}$                &     -                                        &       -                         \\
+ SLLM (Inference)        &  $1/2 \times {T} \times \mathcal{T}_{img} $  &     -                                       &   $1/2 \times {T} \times \mathcal{T}_{res}$ \\
+ SLLM (Training)         &  $1/2 \times{T} \times \mathcal{T}_{img}$    &  $\approx 1/4 \times {T} \times \mathcal{T}_{img}$  &   $1/2 \times {T} \times \mathcal{T}_{res}$     \\ \midrule
\bottomrule
\end{tabular}
}
\label{Tab:Time_com}
\end{table}

\subsection{Analysis on Time Complexity}
\label{Sec:Time}
We take the ActionCLIP as the baseline for this analysis and show the comparison in Table~\ref{Tab:Time_com}, while other methods can be easily deduced. 
Our approach is mostly employed to optimize the computationally expensive image encoding, which accounts for approximately 94\% of the total time consumption.
We set the sampling filter $r$ in Equation~\ref{Eqn:discarding} to 2, i.e., the sampled frames are downsized by half. In next, we mainly focus on two variable components that affect the model efficiency.

\noindent\textbf{Image Encoder.}
The time consumption for a single image can be expressed as
\begin{equation}
    \mathcal{T}_{img} = \mathcal{O}(L_{img} \times (C^2P + CP^2) + CE),
\end{equation}
where $C$ and $P$ denote the number and size of tokens in one image, respectively; $E$ is the output dimension, and $L_{img}$ is the ViT layer number. 
Specifically, the first term $C^2P$ represents the computational cost of the matrix multiplication involved in mapping the tokens to matrices Q, K, and V. 
The second term $CP^2$ corresponds to the multi-head self-attention mechanism. 
Finally, the third term $CE$ involves a linear function that converts the output of self-attention into the frame feature vector of size $\mathbb{R}^{E}$.
Given that the sampled frame set consists of $T$ frames, the overall time required for image encoding in ActionCLIP is $T \times \mathcal{T}{img}$. 
When using a sampling filter $r=2$, our SLLM empirically reduces the computation by half.

\noindent\textbf{FFRes Module.}
The FFRes module utilizes two consecutive frame features of size $\mathbb{R}^{E}$ to restore the discarded frame features. The time cost of the restoration module $\mathcal{T}_{res}$ can be expressed as follows:
\begin{equation}
\mathcal{T}{res} = \mathcal{O}(L_{ffr} \times (2^2E + 2E^2)),
\end{equation}
where $L_{ffr}$ denotes the number of FFRes layers, and we use three in our implementation.

\noindent\textbf{Combination of these two.}
We compare the time complexity of two components, $\mathcal{T}{img}$ and $\mathcal{T}{res}$, based on their estimated costs of 4.41 GFLOPs and 0.035 GFLOPs, respectively. In the original ActionCLIP with 16 sampled frames, image encoding required 70.62 GFLOPs, but our method using SLLM significantly expedites image encoding to 35.31 GFLOPs with the negligible additional computational overhead of FFRes (0.28 GFLOPs). This implies that during the \textbf{\textit{inference}} stage, our approach utilizes only 50\% of the original computational resources, thereby resulting in improved efficiency of the model.
However, it should be noted that our method still requires fixed feature extraction of discarded frames during \textit{\textbf{training}}. Given that computation of both forward and backward can be roughly estimated as twice that of forward, adding the time costs required for the gradient-free feature extraction process, i.e., less than 18 GFLOPs, our method still remains more efficient than ActionCLIP. Detailed comparison results are provided in Section~\ref{Sec:Exp}.

\section{Experiment}
\label{Sec:Exp}
\subsection{Datasets and Baselines}
We used four public datasets to evaluate the effectiveness of our method, including Kinetics-400~\cite{Kinectics}, ActivityNet~\cite{ActNet}, UCF-101~\cite{UCF101}, and HMDB-51~\cite{HMDB}. We then applied our method to three popular baselines, i.e., TAM~\cite{TAM}, ActionCLIP~\cite{ActionClip}, and Text2Vis~\cite{Transfer_VL_ko}. Among them, TAM is a temporal CNN model with ResNet~\cite{ResNEt} as the backbone and trained from scratch. The latter two are state-of-the-art action recognition models based on vision-language models, which are pre-trained on WIT-400M~\cite{CLIP}.

\begin{table*}[]
\centering
\caption{Comparison between three baselines and ones with the integration of our SLLM method. We report the efficiency metrics on Training and Inference video throughput (video/s). The evaluation of action recognition is measured by Top-1 and Top-5 accuracy (\%).  We highlighted the best performance in bold, and the second-best one is indicated with underlining. $\#T$: the number of sampled frames.}
\scalebox{0.70}{
\begin{tabular}{l|c|cc|cc|cc|cc|cccccccc}
\toprule \midrule
\multicolumn{1}{l|}{\multirow{2}{*}{Model}}     & \multirow{2}{*}{\#$T$} & \multicolumn{2}{c|}{Encoder} & \multicolumn{2}{c|}{Efficiency} & \multicolumn{2}{c|}{$\Delta$ Efficiency} & \multicolumn{2}{c|}{$\Delta$ ACC} & \multicolumn{2}{c}{Kinetics-400} & \multicolumn{2}{c}{ActivityNet} & \multicolumn{2}{c}{UCF-101} & \multicolumn{2}{c}{HMDB-51} \\ \cmidrule(lr){3-4} \cmidrule(lr){5-6} \cmidrule(lr){7-8} \cmidrule(lr){9-10} \cmidrule(lr){11-12} \cmidrule(lr){13-14} \cmidrule(lr){15-16} \cmidrule(lr){17-18}
          &                          & Vis  & \multicolumn{1}{c|}{Txt}  & Train.  & Infer. & Train. & Infer.  & Top-1  & Top-5 & Top-1       & Top-5      & Top-1        & Top-5        & Top-1          & Top-5          & Top-1        & Top-5       \\ \midrule
TAM       & 8            & ResNet   & $\times$        & 30.45   & 85.61     &  -    &   -    &  -   &  -    & 56.17             & 80.96              & 56.61            & 80.95          &99.92           & 99.95          & 73.81        &91.55  \\
TAM       & 16           & ResNet   & $\times$        & 15.27   & 42.26     &  -    &   -    &  -   &  -    & 57.86             & 85.15              & 59.68            & 82.56          &99.97           & 99.99          & 75.11        &92.20 \\
TAM+SLLM & 8            & ResNet   & $\times$        & 37.54   & 151.39    &+23.28 &\underline{+76.84}  &\textbf{-0.22} &-0.45   & 57.32             & 81.02              & 55.31          & 80.34        &99.91           & 99.95          & 73.11        &90.32 \\
TAM+SLLM & 16           & ResNet   & $\times$        & 18.85   & 66.54     &+23.44 &+57.45  &\underline{-0.29} &-0.41   & 57.81             & 84.89              & 58.80          & 81.57        &99.96           & 99.99          & 74.88        &91.79 \\ \midrule 
ActionCLIP       & 8     & ViT-B/32   & \checkmark    & 98.92   & 139.52    &   -   &  -     & -    & -     & 73.98             & 92.92              & 82.33          & 96.53        & 99.95          & 99.98          & 79.36        & 95.65      \\
ActionCLIP       & 16    & ViT-B/32   & \checkmark    & 41.79   & 101.54    &   -   &  -     & -    & -     & 75.72             & 93.54              & 84.69          & \underline{97.52}        & 99.96          & 99.99          & 81.62        & 96.21      \\
ActionCLIP+SLLM & 8     & ViT-B/32   & \checkmark    & \underline{130.24}  & \underline{224.17}    &+31.66 &+60.67 &-0.45 &\underline{-0.05}   & 73.84              & 92.80             & 81.28           & 96.17        & 99.91          & 99.99          & 78.80        & 95.91     \\
ActionCLIP+SLLM & 16    & ViT-B/32   & \checkmark    & 60.18   & 188.15    &+44.01 &\textbf{+85.30}  &-0.38 &-0.17  & 75.23               & 93.42            & 84.25           & 97.03        & 99.97          & 99.99          & 81.03        & 96.13  \\\midrule 
ActionCLIP       & 8     & ViT-B/16   & \checkmark    & 44.26   & 98.78     & -     &   -    & -    & -     & 75.15              & 94.11            & 85.64         & 97.49        & 99.95          & 99.99          & 81.22        & \textbf{97.25}       \\
ActionCLIP       & 16    & ViT-B/16   & \checkmark    & 20.77   & 53.52     & -     &   -    & -    & -     & 76.03              & \textbf{94.54}   & \textbf{87.86}       & \textbf{97.71}        & 99.97          & 99.99          & \textbf{82.93}        & \underline{97.17}       \\
ActionCLIP+SLLM & 8     & ViT-B/16   & \checkmark    & 70.32   & 153.83    &\textbf{+58.88} &+55.72  &-0.44 &-0.50  & 74.27               & 93.27             & 84.90        & 97.16        & 99.95          & 99.99          & 81.10        & 96.42       \\
ActionCLIP+SLLM & 16    & ViT-B/16   & \checkmark    & 30.19   & 86.82     &\underline{+45.35} &+62.22  &-0.48 &-0.37  & 75.78               & \underline{94.27}  & \underline{86.62}        & 97.46        & 99.94          & 99.99          & 82.51        & 96.21       \\ \midrule
Text2Vis      & 8        & ViT-B/32   & \checkmark    & 116.90  & 152.53    &  -    &    -   & -    & -     & 75.02              & 92.27             & 81.69        & 96.45        & 99.98          & 99.98          & 80.40        & 95.35      \\
Text2Vis       & 16      & ViT-B/32   & \checkmark    & 62.85   & 79.21     &  -    &   -    & -    & -     & \textbf{77.91}     & 93.49             & 84.53        & 96.27        & \underline{99.99}          & \underline{99.99}          & \underline{82.74}        & 95.31       \\
Text2Vis+SLLM & 8       & ViT-B/32   & \checkmark    & \textbf{147.38}  & \textbf{232.93}    &+26.07 &+52.71  &-0.43 &-0.40  & 74.47 & 91.83       & 81.03             & 95.63        & 99.97          & 99.98          & 79.91        & 95.01       \\
Text2Vis+SLLM & 16      & ViT-B/32   & \checkmark    & 78.52   & 124.80    &+24.93 &+57.56  &-0.50 &\textbf{-0.02}  & \underline{77.45}   & 93.30             & 84.16        & 96.23        & \textbf{99.99}          & \textbf{99.99}          & 81.55        & 95.46 \\\midrule 
\bottomrule
\end{tabular}
}
\label{Tab:with_backbone}
\end{table*}

\subsection{Implementation Details}
We uniformly sampled 8 or 16 frames for each video clip following the sparse sampling strategy~\cite{TSN}. 
All sampled frames in training and testing sets are resized to $224 \times 224$, followed by image augmentation methods involving random scaling, cropping, and rotation. 
For methods with ViT backbone, we partitioned the images into patches of size $32 \times 32$ or $16 \times 16$ based on backbone architectures, which are subsequently fed into the visual encoder. And the output of this image encoding is a 512-dimensional feature vector.
Pertaining to the FFRes module, the values of filter $r$ used in Equation~\ref{Eqn:discarding} is set to 2, i.e., we discarded the sample frame set by half\footnote{For discarded frames that may not have a following remaining frame, we excluded them from the FFRes module.}.
To align the text processing methods of ActionCLIP and Text2Vis, we fixed the length of action label prompts and captions to 77. 
We applied the traditional prompt, which involves a fixed set of templates, such as `a video of [label],' to baseline models. In contrast, SLLM employs the action label augmentation to generate captions via GPT4~\cite{ChatGPT}.
The $\tau$ in Equation~\ref{Eqn:aug} is set to 0.0 to perform greedy decoding. We employed a transformer-based textual encoder~\cite{CLIP} to map the texts into 512-dimensional vectors.
All experiments are conducted on an RTX 3090 GPU.
We applied identical hyperparameter settings, e.g., batch size and learning rate, to models using the same backbone.



\subsection{Experimental Results}
\label{Sec:res}

\subsubsection{Overall Performance Comparison} 
\label{Sec:Acc and Eff}
We first validated the effectiveness of SLLM by applying it to the three baselines and reported the results in Table~\ref{Tab:with_backbone}. We measured the performance of models from two perspectives. 

\noindent\textbf{Efficiency.} 
We first compared the computation efficiency regarding the video processing speed (video/s), i.e., the throughput of videos, during both training and inference phases. Moreover, we calculated the percentage increase at the same sampling rate in the $\Delta$~Efficiency columns. Specifically, $\Delta$ Efficiency can be formulated as,
\begin{equation}
    \Delta \operatorname{Efficiency} = \frac{V_{res} - V_{ori}}{V_{ori}},
\end{equation}
where $V_{ori}$ and $V_{res}$ are the video throughput of the baselines and the ones integrated SLLM method, respectively.

As seen from Table~\ref{Tab:with_backbone}, our method has achieved a significant improvement in efficiency across all baselines. For instance, the inference throughput of ActionCLIP has increased by over 60\% when employing the ViT-B/32 as the visual encoder. And the Text2Vis process more than 230 videos per second when sampling eight frames. Another notable result is that the acceleration achieved during the training phase is inferior to that in the inference. This discrepancy can be attributed to the additional time cost incurred by the fixed vision feature extraction in training. 

\begin{table}[]
\caption{Results of the ActionCLIP's average accuracy across four datasets and GFLOPs. 
$\#T$: the number of sampled frames, 
$\#T_{img}$: the number of frames inputted to the vision encoder,
$\#T_{res}$: the number of features restored by the FFRes module.}
\centering
\scalebox{0.70}{
\begin{tabular}{lcccc|cc}
\toprule \midrule
\multicolumn{1}{l|}{Model}            & $\#T$   & $\#T_{img}$  & $\#T_{res}$    & GFLOPs   & Top-1        & Top-5   \\    \midrule    
\multicolumn{1}{l|}{ActionCLIP}       & 8   &     8            &  0             & 25.61   & 83.91      & 96.27     \\  \midrule
\rowcolor[HTML]{F5F4E9}\multicolumn{1}{l|}{ActionCLIP+SLLM}     & 16   &  8  & 8  & 25.88   & 85.12    & 96.64   \\       
\multicolumn{1}{l|}{ActionCLIP}       & 16    &  16           & 0                   &  49.29    & 85.50    & 96.82      \\ \midrule
\rowcolor[HTML]{F5F4E9}\multicolumn{1}{l|}{ActionCLIP+SLLM}  & 32     & 16   & 16    & 49.85  & 85.70  & 96.95   \\        
\multicolumn{1}{l|}{ActionCLIP}       & 32    &   32        &    0             & 96.63    & 86.00        & 97.03     \\ \midrule 
\bottomrule
\end{tabular}
}
\label{Tab:Same_baselines}
\end{table}

\noindent\textbf{Accuracy.}
Besides the standard Top-1 and Top-5 accuracy metrics, we also calculated the $\Delta$ ACC to measure the performance change,
\begin{equation}
    \Delta \operatorname{ACC} = A_{res} - A_{ori},
\end{equation}
where $A_{ori}$ and $A_{res}$ are the average accuracy over four used datasets of baselines and the ones integrated with SLLM, respectively.
It can be observed that SLLM leads to a slight performance degradation. For instance, TAM exhibits a marginal performance drop of 0.22\% and 0.29\% in sampling 8 and 16 frames, respectively, whereas Text2Vis demonstrates a minimal decrease of only 0.02\% in Top-5 accuracy. These results indicate that our SLLM improves efficiency while maintaining favorable model performance.

\subsubsection{Comparison on Different Sample Frames} 
We also perform comparisons across different numbers of sampled frames and show the results in Table~\ref{Tab:Same_baselines}.
It can be observed that our method gains a better efficiency-effectiveness trade-off than the ActionCLIP baseline. 
Specifically, on the one hand, when sampling 16 frames, SLLM employs 25 GFLOPs to achieve the comparable accuracy ($\approx$85\%) that ActionCLIP requires 49 GFLOPs to attain. 
On the other hand, when using the same computational resources, SLLM can obtain a significant improvement in efficiency. For example, for the baseline with and without our method using eight sampled frames, the averaged accuracy is 85\% v.s. 83\%.

\begin{table}[]
\caption{Model performance under the zero-shot setting. All the models are trained on the Kinetics-400 dataset and tested on the other three ones. $\#T$: the number of sampled frames.}
\centering
\scalebox{0.70}{
\begin{tabular}{l|c|cccccc}
\toprule \midrule
\multicolumn{1}{l|}{\multirow{2}{*}{Model}}   & \multirow{2}{*}{$\#T$}  & \multicolumn{2}{c}{ActivityNet}    & \multicolumn{2}{c}{UCF-101}  & \multicolumn{2}{c}{HMDB-51} \\ \cmidrule(lr){3-4} \cmidrule(lr){5-6} \cmidrule(lr){7-8} 
      & & Top-1 & \multicolumn{1}{c}{Top-5} & Top-1 & \multicolumn{1}{c}{Top-5} & Top-1 &Top-5 \\    \midrule
\multicolumn{1}{l|}{ActionCLIP}                                       & 8   &78.26 &95.50 & 70.75  &91.51   & 48.80 & 72.54    \\ 
\multicolumn{1}{l|}{ActionCLIP}                                       & 16  &78.10 &95.71 & 70.44  &91.19   & \textbf{48.96} & 73.81           \\ 
\rowcolor[HTML]{F5F4E9}\multicolumn{1}{l|}{ActionCLIP+SLLM}          & 8   &78.10 &96.82 & 72.93  &\textbf{96.37}   & 44.41 & 71.13       \\   
\rowcolor[HTML]{F5F4E9}\multicolumn{1}{l|}{ActionCLIP+SLLM}          & 16  &\textbf{80.07} &\textbf{97.19} & \textbf{73.50}  &95.99   & 46.28 & \textbf{74.33} \\ \midrule    
\multicolumn{1}{l|}{Text2Vis}                                         & 8   &72.21 &92.57 & 66.77  & 87.01  & 35.26 & 62.87    \\ 
\multicolumn{1}{l|}{Text2Vis}                                         & 16  &73.67 &92.99 & 66.41  & 87.65  & 36.01 & 64.36     \\ 
\rowcolor[HTML]{F5F4E9}\multicolumn{1}{l|}{Text2Vis+SLLM}            & 8   &74.29 &93.62 & 72.26  & 93.31  & \textbf{42.26} & 68.31 \\   
\rowcolor[HTML]{F5F4E9}\multicolumn{1}{l|}{Text2Vis+SLLM}            & 16  &\textbf{77.27} &\textbf{95.25} & \textbf{74.19}  & \textbf{94.05}  & 41.89 & \textbf{69.05}      \\ \midrule    
\bottomrule
\end{tabular}
}
\label{Tab:Zero_Shot}
\end{table}

\subsubsection{Zero-shot Performance} 

To further verify the effectiveness of our SLLM module under the zero-shot setting, we trained ActionCLIP (with ViT-B/16 as the backbone) and Text2Vis on kinetics-400 and tested them directly on the other three datasets. 
A surprising observation can be found in Table~\ref{Tab:Zero_Shot}: SLLM greatly enhances the generalization capability of the models, as opposed to its slight performance degradation in the original setting. 
For instance, ActionCLIP achieves approximately 2\% increased Top-1 accuracy on ActivityNet, while Text2Vis gains over 5\% improvement on HMDB-51. There are two possible reasons for this result: 
i) The pre-trained captioner makes the action labels more distinguishable, and this ability is further enhanced with unseen actions;
ii) The frame discarding and feature restoration can be regarded as a regularization for the original objective, and thus alleviating over-fitting. 

\begin{table}
\centering
\caption{Performance of SOTAs on Kinetics-400 dataset. We partitioned the model into two groups based on the number of sampled frames. $\#T$: the number of sampled frames.}
\scalebox{0.80}{
\begin{tabular}{l|c|cc}
\toprule
\midrule
Model                    &\#$T$              & Top-1   & Top-5              \\ \midrule
I3D-50~\cite{I3D}                   &  8                & 70.60           & 90.58               \\ 
I3D-101~\cite{I3D}                  &  8                & 71.94           & 91.56               \\
ViViT-B/32\_S~\cite{vivit}             &  8                & 74.23           & 92.25               \\
TimeSformer\_S~\cite{Timesformer}            &  8                & 72.01           & 90.75               \\
\rowcolor[HTML]{F5F4E9}ActionCLIP + SLLM       &  8                & 73.84           & \textbf{92.80}      \\ 
\rowcolor[HTML]{F5F4E9}Text2Vis + SLLM         &  8                & \textbf{74.47}  & 91.83               \\ \midrule
SlowFast~\cite{SlowFast}                       & 16+64             & 77.00           & 92.60             \\
S3D~\cite{S3D}                          & 64                & 74.70           & 93.40            \\
R(2+1)D~\cite{R2+1D}                        & 64                & 72.00           & 90.00            \\
ViViT-B/32\_M~\cite{vivit}                    &  16               & 74.65           & 92.80               \\
ViViT-B/32\_L~\cite{vivit}                   &  64               & 76.78           & 94.26              \\
TimeSformer\_L~\cite{Timesformer}                  &  96               & \textbf{79.65}           & 93.75                \\ 
\rowcolor[HTML]{F5F4E9}Text2Vis + SLLM          &  16                & 77.45         & 93.30               \\ 
\rowcolor[HTML]{F5F4E9}ActionCLIP + SLLM        &  16                & 75.78         & \textbf{94.27}               \\ \midrule
\bottomrule
\end{tabular}
}
\label{Tab:with_SOTA}
\end{table}

\subsubsection{Comparison with SOTAs}
The results of our SLLM methods and several SOTA recognition models are presented in Table~\ref{Tab:with_SOTA}. Pertaining to the first group, our SLLM module demonstrates the best performance among several strong baselines.
Our method outperforms TimeSformer by 2.46\% and 1.08\% in terms of Top-1 and Top-5, respectively.
As for the second group, our method performs comparably with the baselines. For instance, the Text2Vis + SLLM achieves over 77\% Top-1 accuracy, better than the baselines with 16 sampled frames, and exhibits competitive performance with several SOTAs using 32 or more frames, such as S3D and SlowFast.

\subsection{Ablation Studies}
\label{Sec:abl}


\begin{table}[]
\caption{Ablated results on SLLM components. We employed ActionCLIP as the baseline. $\#T_{img}$: the number of frames inputted to the vision encoder,
$\#T_{res}$: the number of features restored by the FFRes module.}
\centering
\scalebox{0.70}{
\begin{tabular}{lcccccc}
\toprule \midrule
\multicolumn{1}{l|}{Model}              & $\#T_{img}$  & $\#T_{res}$  & GFLOPs   & Kinetics-400        & UCF-101    \\    \midrule
\rowcolor[HTML]{F5F4E9}\multicolumn{1}{l|}{ActionCLIP+SLLM}         & 8 & 8 & 38.52   & 75.23     & 73.50      \\  \midrule
\multicolumn{1}{l|}{\quad \textit{w/o} Supervision}                   &  8 & 8 & 38.52  & 72.89 (\textcolor{blue}{-2.34})    & 73.18 (\textcolor{blue}{-0.32})      \\ 
\multicolumn{1}{l|}{\quad \textit{w/o} FFRes}                         &  8 & 0 & 38.41   & 73.83 (\textcolor{blue}{-1.40})    & 73.21 (\textcolor{blue}{-0.29})    \\ 
\multicolumn{1}{l|}{\quad \textit{w/o} Augmentation}                  & 8 & 8 & 38.52  & 74.59 (\textcolor{blue}{-0.64})  & 70.84 (\textcolor{blue}{-2.66})     \\  
\multicolumn{1}{l|}{\quad \textit{w/o} Both}             & 8 & 0 & 38.41  & 73.98 (\textcolor{blue}{-1.25})    & 70.75 (\textcolor{blue}{-2.75})    \\  \midrule
\bottomrule
\end{tabular}
}
\label{Tab:abl_gene}
\end{table}

\begin{table}[]
\caption{Cosine similarity between restored features and discarded ones. We employed ActionCLIP as the baseline. $\#T$: the number of sampled frames,
$\#T_{res}$: the number of features restored by the FFRes module.}
\centering
\scalebox{0.70}{
\begin{tabular}{lc|cc|cc}
\toprule \midrule
\multicolumn{1}{l|}{Model}      &Vis            & $\#T$    & $\#T_{res}$    & Similarity   & Kinetics-400      \\    \midrule    
\rowcolor[HTML]{F5F4E9}\multicolumn{1}{l|}{ActionCLIP+SLLM}       &ViT-B/32       & 8       & 4   & 0.8190   &     73.84       \\  
\rowcolor[HTML]{F5F4E9}\multicolumn{1}{l|}{ActionCLIP+SLLM}       &ViT-B/32       & 16      & 8   & 0.8176   &     75.23   \\       
\rowcolor[HTML]{F5F4E9}\multicolumn{1}{l|}{ActionCLIP+SLLM}       &ViT-B/16       & 8       & 4   & 0.8208   &     74.27   \\       
\rowcolor[HTML]{F5F4E9}\multicolumn{1}{l|}{ActionCLIP+SLLM}       &ViT-B/16       & 16      & 8   & 0.8218   &     75.78   \\       \midrule
\multicolumn{1}{l|}{\quad\hspace{1pt}\textit{w/o} Supervision}       &ViT-B/32       & 8       & 4   & 0.6324    & 70.97          \\ 
\multicolumn{1}{l|}{\quad\hspace{1pt}\textit{w/o} Supervision}       &ViT-B/32       & 16      & 8   & 0.6385  & 72.89     \\      
\multicolumn{1}{l|}{\quad\hspace{1pt}\textit{w/o} Supervision}       &ViT-B/16       & 8       & 4   & 0.6517   & 73.14       \\       
\multicolumn{1}{l|}{\quad\hspace{1pt}\textit{w/o} Supervision}       &ViT-B/16       & 16      & 8   & 0.6682   & 74.40       \\       \midrule
\bottomrule
\end{tabular}
}
\label{Tab:Cos_simi}
\vspace{-2pt}
\end{table}

\subsubsection{Performance of Several Model Variants}
We conducted ablation studies to investigate the impact of removing specific components in SLLM. To this end, we designed four ablated variants with certain components removal as follows: i) The supervision process from discarded frames during training; ii) The full FFRes module but with the label augmentation, i.e., the baselines are trained with downsized sampled frames without restoration; iii) The action label augmentation; iv) The complete SLLM, namely, the model degrades to an ActionCLIP with low-sampled frames. We summarized these results in Table~\ref{Tab:abl_gene}. These models are trained on the Kinetics-400 dataset and tested on the UCF-101 dataset under a zero-shot setting.

Our observations consist of two primary findings: 
i) Removing any single component results in a notable decline in the overall performance. The FFRes module contributes significantly to the improvements achieved, as evidenced by a 2\% reduction in performance when removed; 
ii) Compared to the FFRes module, label augmentation serves as the primary driving force for improving the generalization capabilities. It is worth noting that the discarding/restoration process does not negatively impact the generalization ability. For instance, the zero-shot performance of the FFRes-only variant is nearly identical to that of the ActionCLIP model.


\begin{figure}[t]
    \centering
    \includegraphics[width=0.42\textwidth]{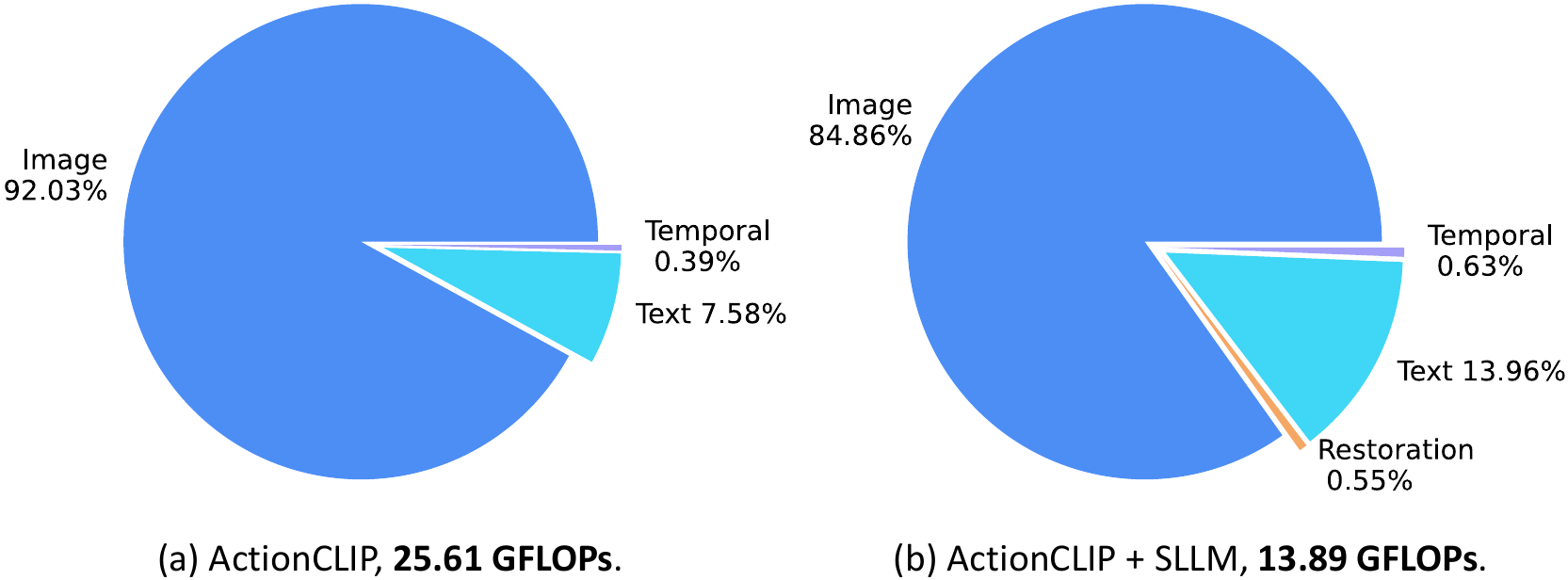}
    \caption{Illustration for time consumption of different components in ActionCLIP with eight frames sampled. With the SLLM integrated, the computational burden of baseline downsizes by about 50\%.}
    \label{fig:pietime}
\end{figure}

\subsubsection{Cosine Similarity of Restored Features}
We further investigated the effectiveness of SLLM in restoring discarded features. Specifically, we calculated the cosine similarity between the restored features and the original discarded frame features and reported the results in Table~\ref{Tab:Cos_simi}. Specifically, we compared the complete SLLM with its ablation version that removed the supervisory signal. The experimental results suggest that the SLLM was able to restore the discarded frame features to a great extent, thereby contributing to the maintenance of the models' performance. Nonetheless, it is worth noting that a poorly designed restoration module can potentially compromise action recognition capability.

\subsubsection{Time Consumption of Components}
\label{Sec:time_comp}
We provide a fractional time consumption of ActionCLIP's various components in Figure~\ref{fig:pietime}a. As analyzed in Section~\ref{Sec:Time}, our SLLM primarily focuses on optimizing the resource-intensive image encoding process. Notably, we observe that the computational overhead of image encoding decreased significantly by more than 50\% from 23.57 GFLOPs to 11.79 GFLOPs, as illustrated in Figure~\ref{fig:pietime}b. This reduction in computational overhead has a corresponding effect on the overall resource consumption of the model, improving the efficacy of our SLLM approach in optimizing the efficiency of the ActionCLIP model.

\subsubsection{Performance with Respect to Sampling Filter}
\label{Sec:FFRES_ABL}
\begin{table}[]
\caption{The accuracy and computational requirements of SLLM with various sampling filter values. Models are trained on the Kinetics-400 and tested on the other three datasets under the zero-shot setting. $\#T_{img}$: the number of frames inputted to the vision encoder.}
\vspace{-0.5em}
\centering
\scalebox{0.70}{
\begin{tabular}{l|c|cc|ccc}
\toprule \midrule
\multicolumn{1}{l|}{Model}    & $\#T_{img}$  & GFLOPs   &K-400     & ActNet   & UCF  &HMDB   \\    \midrule    
\multicolumn{1}{l|}{ActionCLIP}                       & 16    & 73.87    & 75.72   & 78.10    & 70.44 & 48.96 \\  \midrule  
\multicolumn{1}{l|}{ + SLLM \textit{w} $r$=2}           & 8     & 38.52    & 75.23   & 80.70    & 73.50 & 46.28 \\ 
\multicolumn{1}{l|}{ + SLLM \textit{w} $r$=3}            & 6     & 29.62    & 73.76   & 76.81    & 73.67 &    44.49 \\        
\multicolumn{1}{l|}{ + SLLM \textit{w} $r$=4}            & 4     & 20.96    & 71.96   & 75.90    & 72.64 & 43.60       \\ \midrule 
\bottomrule
\end{tabular}
}
\label{Tab:visual_branch}
\end{table}

The results of the SLLM with varying sampling filters $r$ are presented in Table~\ref{Tab:visual_branch}. It is easy to infer that increasing the value of $r$ leads to a significant reduction in the number of encoded frames $\#T^{\prime}$, which in turn enhances the efficiency of the model. However, this efficiency gain comes at a cost, as the reduction in encoded frames also leads to a noticeable deterioration in restoration effects. As an example, we observe that when $r$ equals 4, the model was only able to achieve 71\% accuracy on the Kinetics-400 dataset compared to 75.72\% in original ActionCLIP. These results highlight the need to balance efficiency gains with model performance carefully.


\subsubsection{Case Study on Label Augmentation}

\begin{table}[]
\centering
\caption{Ten actions that bear the greatest resemblance to the `Dancing macarena' across various prompt formats. Our methodology enhances the discriminability between actions.}
\vspace{-0.5em}
\scalebox{0.70}{
\begin{tabular}{l|lc|lc}
\toprule \midrule
Action               & \multicolumn{4}{c}{Dancing macarena}     \\  \midrule 
\multirow{2}{*}{Prompt} & \multicolumn{2}{c|}{Vanilla} & \multicolumn{2}{c}{Augmentation} \\ \cmidrule(lr){2-3} \cmidrule(lr){4-5} 
                        &  \multicolumn{1}{c}{Action}     &  Simi.         & \multicolumn{1}{c}{Action}    & Simi.\\ \midrule
\multirow{10}{*}{Top-10}   &Dancing gangnam style  &0.8989  & Dancing gangnam style   & 0.7871             \\
                           &Dancing ballet         &0.8862   &Country line dancing     & 0.7856            \\
                           &Dancing charleston     &0.8657   &Salsa dancing            & 0.7822      \\ 
                           &Doing aerobics         &0.8633   &Clapping                 & 0.7778      \\
                           &Krumping               &0.8594   &Dancing ballet           & 0.7759            \\
                           &Salsa dancing          &0.8574   &Dancing charleston       & 0.7695       \\ 
                           &Tap dancing            &0.8540   &Robot dancing            & 0.7686         \\                
                           &Jumpstyle dancing      &0.8530   &Finger snapping          & 0.7661             \\
                           &Lunge                  &0.8530   &Doing aerobics           & 0.7607       \\ 
                           &Robot dancing          &0.8501   &Jumpstyle dancing        & 0.7559        \\ \midrule                         
\bottomrule
\end{tabular}
}
\vspace{-3pt}
\label{Tab:cos_simi}
\end{table}
Lastly, we investigated the efficacy of our action label augmentation method and compared it with the vanilla prompting approach. Specifically, we employed a pre-trained text extractor~\cite{CLIP} to extract features from the labels applied to vanilla prompts and our augmentation method. 
Table~\ref{Tab:cos_simi} showcases the top 10 actions that bear the closest resemblance to the `dancing macarena' action (excluding itself). Notably, our approach substantially decreases the similarity between actions. 
We further computed the average similarity across the 400 actions in Kinetics-400. When utilizing label augmentation, the similarity value is 0.6610, which is significantly lower than the 0.7149 achieved by vanilla methods. This finding highlights the effectiveness of our approach in the statistical dimension as well.

\subsection{Discussion}

\begin{table}[]
\caption{Efficiency (videos/s) and Performance (\%) of Text2Vis with the intergration of different samplers on HMDB-51 dataset. $\#T$: the number of sampled frames, $\#T_{img}$: the number of frames inputted to the vision encoder.}
\centering
\scalebox{0.70}{
\begin{tabular}{l|cc|cc}
\toprule \midrule
\multicolumn{1}{l|}{Model}         & $\#T$    & $\#T_{img}$    & Efficiency   & Top-1      \\    \midrule    
\rowcolor[HTML]{F5F4E9}\multicolumn{5}{l}{Text2Vis}          \\
{\textit{w} MGSampler}                                       & 16      & 16   & 73.28   &     82.96   \\    
{\textit{w} Vanilla Sampler}                                 & 16      & 16   & 79.21   &     82.74   \\      
{\textit{w} MGSampler}                                       & 8       & 8   & 149.10   &     80.97   \\        
{\textit{w} Vanilla Sampler}                                 & 8       & 8   & 152.53   &     80.40   \\ \midrule
\rowcolor[HTML]{F5F4E9}\multicolumn{5}{l}{Text2Vis + SLLM}         \\
{\textit{w} MGSampler}                                       & 16      & 8   & 121.80   &     81.77   \\    
{\textit{w} Vanilla Sampler}                                 & 16      & 8   & 124.80   &     81.55  \\           \midrule  
\bottomrule
\end{tabular}
}
\label{Tab:Sampler}
\vspace{-2pt}
\end{table}

\subsubsection{Model Performance with Samplers}
Existing sampling strategies~\cite{OCSampler} aim primarily to extract representative frames rather than random sampling frames. These frames are determined by the extracted features of all frames. Although recent sampling methods have reduced the workload of feature extraction for all frames, they still unavoidably introduce some additional computations compared to traditional samplers.

Our method alleviates the burden of feature extraction by reconstructing the frame features rather than encoding them from scratch. In this way, our method helps achieve higher efficiency while maintaining model performance largely. We can sample frames using sampling strategies, and then perform our method to achieve a better balance of efficiency and effectiveness. To validate this, we chose MGSampler~\cite{MGSampler} and integrated it into Text2Vis. From Table~\ref{Tab:Sampler}, we observe that the MGSampler outperforms the vanilla sampler, implying that the sampled frames are more representative and important. However, MGSampler introduces additional computational costs. For instance, when sampling 16 frames, the video throughput of Text2Vis + SLLM decreased by 3 videos per second. With the integration of our SLLM, we can achieve a better trade-off between accuracy and efficiency.

\begin{table}[]
\caption{Efficiency (videos/s) and Performance (\%) of Text2Vis with the frozen encoder and FFRes module on HMDB-51 dataset. $\#T$: the number of sampled frames, $\#T_{img}$: the number of frames inputted to the vision encoder.}
\centering
\scalebox{0.70}{
\begin{tabular}{l|cl|cc}
\toprule \midrule
\multicolumn{1}{l|}{Model}         & $\#T$    & $\#T_{img}$    & Efficiency   & Top-1      \\    \midrule    
Text2Vis                                      & 16      & 16   & 79.21   &     82.74          \\ 
Text2Vis + SLLM (Frozen)         & 16      & 8+8 (frozen)   & 79.78   &     82.16          \\ 
\rowcolor[HTML]{F5F4E9}Text2Vis + SLLM (Restoration)                & 16      & 8   & 124.80   &     81.55          \\ \midrule 
Text2Vis                                       & 8      & 8   & 152.53   &     80.40          \\ 
Text2Vis + SLLM (Frozen)         & 8      & 4+4 (frozen)   & 166.28   &     79.98          \\ 
\rowcolor[HTML]{F5F4E9}Text2Vis + SLLM (Restoration)                & 8      & 4   & 232.93   &     79.91          \\        \midrule  
\bottomrule
\end{tabular}
}
\label{Tab:Frozen}
\vspace{-2pt}
\end{table}

\subsubsection{Impact of the Frozen Encoder}
We conducted the experiments on using frozen encoder only. Specifically, we eliminated the feature restoration and instead utilized the frozen supervisions for action recognition. The video throughput and accuracy are summarized in Table~\ref{Tab:Frozen}. As observed in our experiments, this approach leads to a slight performance degradation compared with the base model. During inference, the efficiency is similar to that of the base model, while largely inferior to that of our proposed method.

\begin{table}[]
\caption{Performance (\%) of Text2Vis with different number of restored frames on HMDB-51 dataset. $\#T$: the number of sampled frames, $\#T_{img}$: the number of frames inputted to the vision encoder, $\#T_{res}$: the number of restored frames between two consecutive frames.}
\centering
\scalebox{0.70}{
\begin{tabular}{l|ccc|c}
\toprule
\midrule
Text2Vis+SLLM                             & $\#T$                & $\#T_{img}$          & $\#T_{res}$            & Top-1             \\ \midrule
\multicolumn{1}{l|}{\multirow{2}{*}{r=2}} & 16                   & 8                    & \multicolumn{1}{c|}{0} & 80.40                \\
\multicolumn{1}{l|}{}                     & 16                   & 8                    & \multicolumn{1}{c|}{1} & 81.55                \\ \midrule

\multicolumn{1}{l|}{\multirow{2}{*}{r=3}} & 16                   & 8                    & \multicolumn{1}{c|}{0} & 78.79                \\
\multicolumn{1}{l|}{}                     & 16                   & 8                    & \multicolumn{1}{c|}{1} & 80.39                \\
\multicolumn{1}{l|}{}                     & 16                   & 8                    & \multicolumn{1}{c|}{2} & 80.20                \\\midrule

\multicolumn{1}{l|}{\multirow{2}{*}{r=4}} & 16                   & 8                    & \multicolumn{1}{c|}{0} & 78.19                \\
\multicolumn{1}{l|}{}                     & 16                   & 8                    & \multicolumn{1}{c|}{1} & 78.86                \\
\multicolumn{1}{l|}{}                     & 16                   & 8                    & \multicolumn{1}{c|}{3} & 78.71                \\ \midrule \bottomrule
                        
\end{tabular}}
\label{Tab:Restored_Settings}
\end{table}

\subsubsection{Discussion on Restoration Settings}
We conducted experiments to evaluate the model under different number of restored frames. In particular, we simply restored one frame feature when $r>2$.
The summarized results are presented in Table~\ref{Tab:Restored_Settings}. In this table, $\#T_{res}$ represents the number of restored frames between two sampled adjacent frames. For instance, $\#T_{res}$ = 1 corresponds to the restoration of a single frame feature with $\lambda$=0.5. From the table, we have three important observations: 1) The model performance gradually decreases as the number of discarded frames increases. 2) Our restoration approach significantly enhances the model performance, as seen for all cases when $\#T_{res}$ > 0 v.s. $\#T_{res}$ = 0. 3) Restoring one frame often slightly outperforms other settings with more frames, which might be because feature resotration is a linear interpolation process. For example, when $r$=4, there are three intermediate frame features to be restored. There exists linear relationship between these three restored features but non-linear relationship between the restored and original features. This incoherency may impair subsequent temporal modeling. 

One potential future direction for this problem is to introduce learnable temporal position embedding to guide the feature restoration of different intermediate frames. As a preliminary exploration, we added temporal positional embeddings by using randomly initialized placeholder tensors as discarded features. The performance of this approach is summarized in Table~\ref{Tab:Tempo}. From these results, we found that the model performance remains similar for $r$ = 2. This can be attributed to the fact the remaining frames are more than other settings, containing essential information to describe an action. However, when we continue to remove more frames, i.e., $r$=3, the temporal information becomes more important and shows a large improvement over the model without the temporal modeling. Lastly, in the context of $r$=4, the randomly initialized placeholder of the discarded frames hardly learns a meaningful temporal embedding, leading to a minor deterioration in model performance.

\begin{table}[]
\caption{Performance (\%) of Text2Vis with restoration approach of linear and temporal embeddings on HMDB-51 dataset. $\#T$: the number of sampled frames, $\#T_{img}$: the number of frames inputted to the vision encoder.}
\centering
\scalebox{0.70}{
\begin{tabular}{l|cc|cc}
\toprule \midrule
Text2Vis                                     & $\#T$                & $\#T_{img}$          & Linear    & Temporal             \\ \midrule
\multicolumn{1}{l|}{r=2}                     & 16                & 8            & 81.55    & 81.47            \\
\multicolumn{1}{l|}{r=3}                     & 16                   & 6         & 80.20    & 81.45            \\ 
\multicolumn{1}{l|}{r=4}                     & 16                   & 4         & 78.71    & 78.57           \\ \midrule \bottomrule     
\end{tabular}}
\label{Tab:Tempo}
\end{table}

\section{Conclusion}
In this paper, we introduce a novel feature restoration approach to enhance the efficiency of video action recognition. Our method alleviates the computational burden associated with image encoding by reducing the number of sampled frames.
We evaluate the generalization capability of this method by applying it to three recent, robust baselines. Experimental results on four widely-used datasets consistently demonstrate its substantial improvements in efficiency while maintaining comparable model performance.

In summary, our proposed approach highlights the potential of utilizing fewer frames to address efficiency challenges in action recognition. Beyond its promising application across various vision backbones, this method also facilitates scalability in terms of model size and training batch size. However, one limitation of our approach is the necessity to extract features from discarded frames for restoration supervision. To overcome this constraint, future research can explore more advanced feature restoration techniques and robust out-of-distribution generation methods.

{\small
\bibliographystyle{ieee_fullname}
\bibliography{egbib}
}

\end{document}